\documentclass[runningheads]{llncs}
\usepackage{makeidx}
\usepackage{graphicx}
\usepackage{stfloats}
\usepackage{epstopdf}
\usepackage{algorithm,algpseudocode}
\usepackage{epsfig}
\usepackage{epstopdf}
\usepackage{amsmath,amssymb} 
\usepackage{color}
\usepackage{listings}
\usepackage{stackengine}
\usepackage{amssymb}
\usepackage{multirow}
\usepackage[inline]{enumitem}
\usepackage{stfloats}

\newcommand{\etal}{\textit{et al}. }
\newcommand{\ie}{\textit{i}.\textit{e}., }

\begin{document}
	\pagestyle{headings}
	\mainmatter

	\def\GCPR16SubNumber{11}

	\title{Automatic Trimap Generation for Image Matting}

	\titlerunning{ }
	\authorrunning{ }
	\author{Vikas Gupta, Shanmuganathan Raman \\ gupta.vikas@iitgn.ac.in, shanmuga@iitgn.ac.in}
	\institute{Electrical Engineering \\ Indian Institute of Technology, Gandhinagar}

	\maketitle

	\begin{abstract}
		
	Image matting is a longstanding problem in computational photography. Although, it has been studied for more than two decades, yet there is a challenge of developing an automatic matting algorithm which does not require any human efforts. Most of the state-of-the-art matting algorithms require human intervention in the form of trimap or scribbles to generate the alpha matte form the input image. In this paper, we present a simple and efficient approach to automatically generate the trimap from the input image and make the whole matting process free from human-in-the-loop. We use learning based matting method to generate the matte from the automatically generated trimap.  Experimental results demonstrate that our method produces good quality trimap which results into accurate matte estimation. We validate our results by replacing the automatically generated trimap by manually created trimap while using the same image matting algorithm. 	
		
	\end{abstract}

	\section{Introduction}
	\label{sec:introduction}
     Image matting is the process of accurately estimating the foreground object in images and videos. It is a very important technique in image and video editing applications, particularly in film production for creating  visual effects. In case of image segmentation, we segment the image into foreground and background by labeling the pixels. Image segmentation generates a binary image, in which a pixel either belongs to foreground or background. However, image matting is different from the image segmentation, wherein some pixels may belong to foreground as well as background, such pixels are called \emph{partial} or \emph{mixed} pixels. Image matting is concerned about determining the convex combination of foreground and background intensity for each partial pixel.
     Porter and Duff in 1984 first proposed the problem of accurately separating a foreground object from the background in order to composite with a new  background for creating a new image, which looks more realistic \cite{porter1984compositing}. The preliminary version of this paper has been published in \cite{gupta2016automatic}.
     \newline  
      Given an image $I_p$  $(p=(x,y))$ , the image matting problem is mathematically stated as given in equation \ref{eq_1}.
     \begin{equation}
     \label{eq_1}
      I_p = \alpha_pF_p + (1-\alpha_p)B_p.
     \end{equation}
      Where, $\alpha_p$  represents the matte and it can take any value in $[0,1]$, and $F_p$ and $B_p$ are foreground and background pixel value respectively. If $\alpha_p = 1 \; \text{or} \; 0$  then  the pixel at location $p$ belongs to  \emph{ definite foreground} or \emph{defiinite background} respectively. Otherwise that pixel is called  a \emph{partial} or a \emph{mixed} pixel. 
      In  natural  images,  majority of  pixels  either belong to definite foreground or definite background region. However, in order to fully separate the  foreground  from the background in an image, accurate  estimation of the alpha values for partial or mixed pixels is necessary.  Note that in   equation \ref{eq_1}, if we consider a full color image (RGB), there are $7 $ unknowns ($F_p, B_p$ for each color channel and $\alpha_p$)  and three equations (one for each color channel). Thus image matting problem is a severely under-constrained problem. Such under-constrained problems can be solved by adding more information into it. This additional information is provided in the form of \emph{trimap} \cite{chuang2001bayesian} or \emph{scribbles} \cite{wang2005iterative}, \ie  labeling some pixels belonging  to definite foreground or definite background.  
      In order to fully extract meaningful foreground object, almost all the matting techniques rely on the user intervention, wherein the user segments the input image into three regions: definite foreground, definite background, and unknown region. This three-level map is called as a \emph{trimap}. Now the matting problem is reduced, and it will have to determine the values of $F_p, B_p$ and $\alpha_p$ for the pixels in the unknown region based on the available information of definite foreground and   definite background region. 
      Instead of carefully labeling the input image into three regions to generate a trimap, some recently proposed methods rely on the user to provide few foreground and background scribbles as input to extract a matte. So this method marks majority of pixels as unknown region. Ideally, the trimap should consist of very small unknown region around the foreground boundary, and it should contain only the partial or mixed pixels. Since smaller  the unknown region (less number of mixed pixels)  the more the accurate will be. However generating such an accurate trimap requires lot of human efforts and it is often undesirable, particularly in the case of transparent objects.
      Thus, accuracy of a trimap is one of the important factors which affects the performance of a matting algorithm \cite{wang2007image}. So, while developing a matting algorithm there will always be a trade off between the accuracy of the matte and the amount of user efforts required. Recently, Levin  \etal proposed spectral matting algorithm \cite{levin2008spectral}, which automatically extracts the matte from the input image without any user intervention. However,  the limitation of this method is that it generates  erroneous result for images with highly- textured background. Therefore, to alleviate such problems user specified \emph{trimap} or \emph{scribbles} are needed to get the highly accurate matte. However, we can reduce the user efforts for manually generating the trimap by automatically generating more accurate trimap. 
      \newline          
      In this paper,  we propose a novel method to automatically generate trimap from the given image. We use the saliency map of the image to generate the trimap. First, we oversegment the image using  SLIC superpixel algorithm \cite{achanta2012slic}. Then we obtain the local features using \emph{Oriented Texture Curves} (OTC) feature descriptor \cite{margolin2014otc} for each superpixel in the over-segmented image. These feature vectors are then clustered to obtain the background and foreground superpixels. Then we update the saliency map of the image and threshold it to obtain the binary map. This binary map is then eroded and dilated in order to obtain the desired trimap.  The steps involved in the proposed method is depicted in Fig. \ref{Fig-1:Flowchart}. The main contributions of our paper are  given below.
      \begin{enumerate}
      	\item  We propose an automatic trimap generation framework for image matting to get rid of any human intervention.
      	\item  Instead of working on each pixel, we employ superpixels  to over-segment the image and process a group of pixels together.
      	\item  We use image saliency and an appropriate local feature
      	descriptor to identify the foreground and background superpixels  which helps in automatic generation of trimap.
      \end{enumerate}
      
       The rest of the paper is organized as follows. In section \ref{sec:related work}, we briefly survey the state-of-the-art matting algorithms as well as existing methods for automatic trimap generation. Section \ref{proposed approach} gives the details of the proposed automatic trimap generation algorithm. In section \ref{Results}, we show and discuss the results of image matting  obtained using the trimap generated from our approach. Section \ref{conclusion} concludes the paper with some ideas for future improvement. 

     \begin{figure}
     \centering
     \includegraphics[width=1\linewidth]{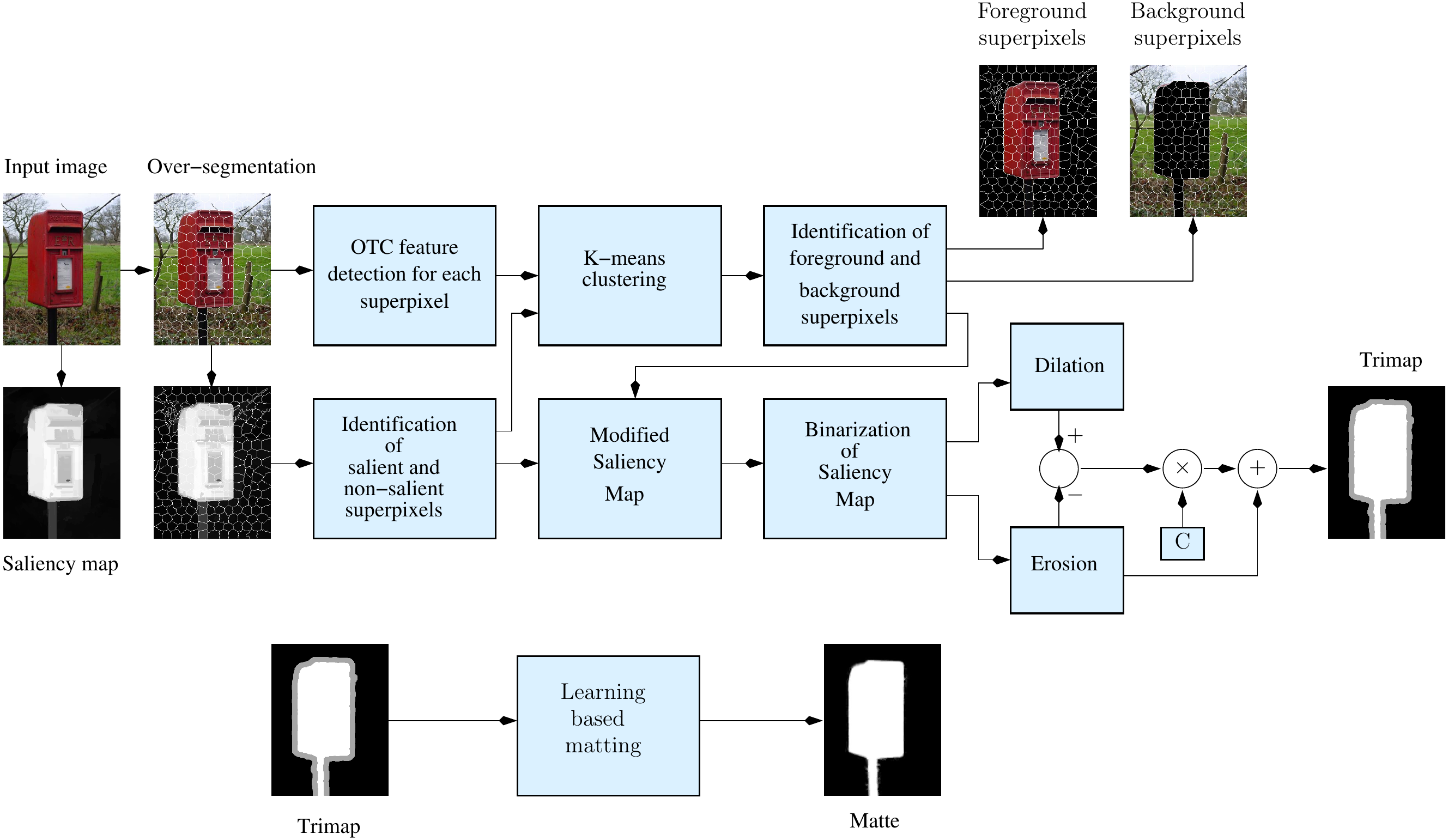}
     \caption{Proposed saliency based automatic trimap generation framework.}
     \label{Fig-1:Flowchart}
     \end{figure}

\section{Related Work}
	\label{sec:related work}
      In this section we review previous work relevant to our work. In particular, we discuss some of the recent state-of-the-art matting algorithms as well as existing methods for automatic trimap generation. Generally the matting algorithms are classified as \emph{Sampling based approaches} \cite{ruzon2000alpha,chuang2001bayesian,wang2005iterative} and \emph{Affinity based approaches} \cite{sun2004poisson,grady2005random,bai2007geodesic,zheng2008fuzzymatte,levin2008closed,levin2008spectral}.

\subsection{Sampling based approaches}
\label{Sampling based matting algo} 
      The basic principle of these approaches is to use neighboring  foreground and background pixels as samples to estimate the alpha values for the unknown pixels.  Ruzon and Tomasi proposed a \emph{sampling} based approach \cite{ruzon2000alpha} for matting. In this approach, alpha values are measured along a manifold connecting the ``frontiers'' of each object's color distribution. The unknown region is divided  into subregions and a local window is defined in these subregions such that it covers the unknown region, and a local foreground and background region. The optimal alpha is the one that yields an intermediate distribution for which the observed color has maximum probability.\\
      The \emph{Bayesian approach} proposed by Chuang \etal   also uses probabilistic approach to solve the matting problem \cite{chuang2001bayesian}. The main difference is that a continuously sliding window is used for selecting the neighborhood, which marches inward from the foreground and background regions. These foreground and background samples are used to build color distributions. The matting problem is formulated in a well-defined Bayesian framework and maximum a posteriori (MAP) technique is used to solve for the matte. \\
      The previous two methods assumes that the unknown region is the narrow band around the foreground boundary and therefore they use local color models. So there are ample amount of foreground and background pixels within a local window centered on any unknown pixel. But this assumption fails if the trimap is not well defined and it  consist of only a few scribbles. In the case of rough trimap, global sampling method is used to tackle the sampling problem. Wang and Cohen proposed an \emph{iterative optimization} based matting approach which computes Gaussian Mixture Models (GMMs) from the  user marked foreground pixels and background pixels, then assign each marked pixel to one Gaussian for further global sampling  \cite{wang2005iterative}. The sampling based approaches works well when the trimap is well defined.
      
\subsection{Affinity based approaches}
\label{affinity based matting algo} 
      The affinity based approaches do not require explicit foreground and background color information to solve the matting problem. These methods utilize the local image statistics by defining various \emph{affinities} between neighboring pixels to model the matte gradient across the image instead of directly estimating the alpha value at each  pixel. \emph{Poisson matting}  estimates the  matte gradient  from the image using boundary information from a user-supplied trimap and then reconstructs the matte by solving Poisson equation \cite{sun2004poisson}. It is based on the assumption that intensity change in the foreground and the background is smooth.  Grady \etal employed \emph{random walk} algorithm  to calculate the final alpha values based on  affinity \cite{grady2005random}. Given a trimap, for each pixel in the unknown region, its alpha value  is set to be the probability that a random walker starting from this pixel location will reach a pixel in the foreground before striking a pixel in the background, when biased to avoid crossing the foreground boundary. The \emph{geodesic} matting method  measures the weighted geodesic distance from the user-provided scribbles to  the pixels in the unknown region (outside of the scribbles) for labeling them as foreground or background  pixel \cite{bai2007geodesic}.\\      
Zheng \etal proposed an \emph{interactive matting} algorithm which is similar to geodesic matting called  \emph{FuzzyMatte} \cite{zheng2008fuzzymatte}. In this method instead of computing geodesic distance, it computes the  \emph{fuzzy connectedness} between the unknown pixel and the known foreground and background pixels. The final alpha value is then calculated using the fuzzy connectedness. The disadvantage of this method is that the fuzzy connectedness is sensitive to image noise which may lead to the misclassification of  pixels in the unknown region. Closed-form matting approach  explicitly derives a cost function from local smoothness assumptions on foreground and background colors and shows that in the resulting expression, it is possible to analytically eliminate the foreground and background colors to obtain a quadratic cost function in alpha \cite{levin2008closed}. This cost function can be solved by a sparse linear system of equations, which yields the globally optimal alpha matte. The affinity used in this approach does not have any global parameters. Instead, it uses  local estimates of mean and variances which leads to  significant improvement in the performance as demonstrated in \cite{levin2008closed}. The \emph{spectral matting} method uses spectral segmentation techniques to obtain basis set of fuzzy matting components from the smallest eigenvectors of the matting Laplacian. These matting  components are used as building blocks to easily construct semantically meaningful foreground mattes  \cite{levin2008spectral}. The practical applications of this approach is limited as the memory consumption  is very high.

\subsection{Other approaches}
\label{Other matting algo} 
 
      Robust matting method  combines the color sampling and affinity  together in a single optimization process  to get more accurate and robust matting solution \cite{wang2007optimized}. It samples the foreground and background colors for unknown pixels and determines the confidence of these samples. The high confidence samples are chosen to contribute to the matting energy function which is minimized by a Random Walk.  Zheng and Kambhamettu utilized semi-supervised learning to solve the digital matting problem which results in  a local learning based matting approach and a global learning based approach \cite{zheng2009learning}. The local learning based matting approach trains a local alpha-color model for each pixel in the image only based on its neighboring pixels which are considered to be most related and suits better than the scribble based matting. The global learning based approach learns the global alpha-color model from  some chosen labeled pixels closer to the unlabeled pixel, and suits better to the case when a trimap is provided and the unknown region is narrow. We use this image matting algorithm to evaluate the effectiveness of the automatic trimaps generated in this work. 
      
      Most of the existing automatic trimap generation algorithms rely on the binary segmentation of the image to get the initial boundary of the foreground object \cite{li2009faceseg} \cite{hsieh2013automatic}\cite{singh2013automatic} \cite{cho2016automatic}. Singh \etal employed Canny edge detection followed by morphological operations (erosion and dilation) to yield the boundary of the foreground \cite{singh2013automatic}. A corrected trimap is then obtained by applying region growing algorithm to the unknown region of the image obtained by dilating the  foreground boundary. Chang-Lin Hsieh \etal proposed an automatic trimap generation method, which generates an initial guess of trimap form the binary segmented image \cite{hsieh2013automatic}. They employed dynamic brush width method to obtain content adaptive trimap from the initial guess of trimap.  Ahmad Al-Kabbany and Eric Dubois employed Gestalt laws of grouping to generate the trimap automatically \cite{Amir2015Gestalt}. Cho Donghyeon \etal utilized depth information and adaptive analysis of color distribution along the foreground boundary of the light field images\cite{cho2016automatic}. 
      
      In our proposed method, instead of using binary segmentation, we use over-segmentation algorithm \cite{achanta2012slic} to get the superpixels of the image. We use these superpixels to roughly decide the foreground and background region of the image. The detailed process of  automatically generating the trimap is described in the next section.
      
\section{Automatic Trimap Generation}
\label{proposed approach}
     In this section, we describe in detail our proposed framework for automatically generating the \emph{trimap} from a given image. We assume that there is a single salient object present in the given scene. The complete framework is divided into three parts as: over-segmentation and feature description, identification of background and foreground superpixels, and trimap generation and matting.
    
\subsection{ Over-segmentation and Invariant Feature Description}
\begin{figure}
	\centering
	\includegraphics[width=0.4\linewidth]{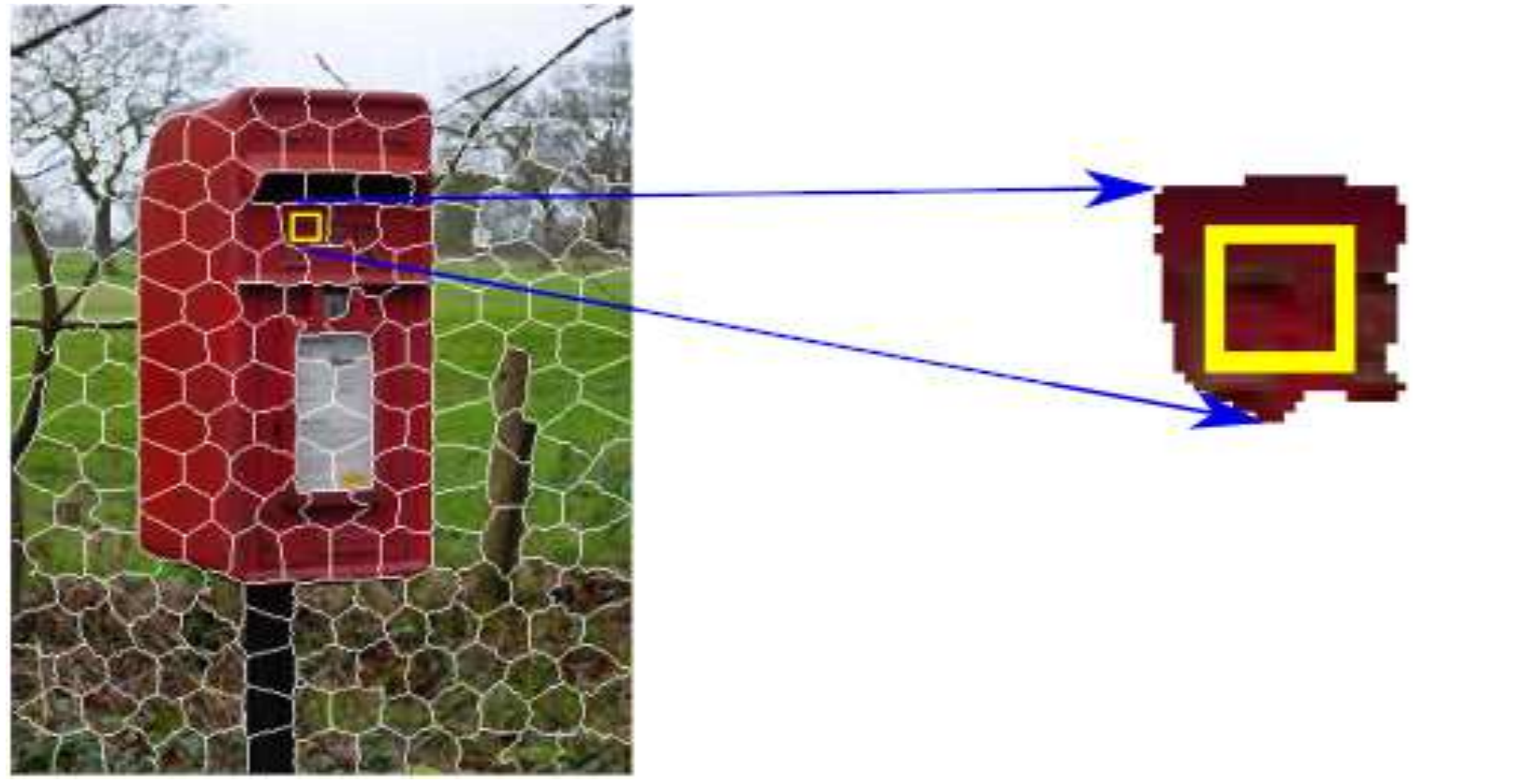}
	\caption{Patch extraction from superpixels: A patch of $13\times13$ is extracted form the superpixels to obtain the OTC features}
	\label{Fig-3: Patch Extraction}
\end{figure}

     Consider an input image $I$ as shown in Fig. \ref{Fig-2: Intermideate Results}(a). At first we segment the  image $I$ into $N$ superpixels using the algorithm given in \cite{achanta2012slic}. Due to $N$ superpixels, the resulting over-segmented image  is shown in Fig. \ref{Fig-2: Intermideate Results}(b). Note that each superpixel contains distinct texture and color information, therefore we compute the OTC features for a patch of size $13\times13$ (see Fig. \ref{Fig-3: Patch Extraction}) in each superpixel \cite{margolin2014otc}. The OTC descriptor captures the texture of a patch along multiple orientations, while maintaining robustness to illumination changes, geometric distortions, and local contrast differences. It provides  a 185-dimensional  texture feature in eight different directions.\\
We obtain the saliency maps $\text{SM}_i$, of the input image using three different methods \cite{jiang2013salient,li2014secrets,zhao2015saliency}. Each of these methods uses different framework to obtain the saliency map. In \cite{jiang2013salient}, Huaizu Jiang \etal employed supervised learning approach to integrate regional features such as the regional contrast, regional property, and regional backgroundness descriptors together to form the master saliency map. In \cite{li2014secrets}, the image is segmented to obtain a set of object candidates and then a fixation algorithm is used to rank different regions based on their saliency score. In \cite{zhao2015saliency}, Rui Zhao \etal utilized global context and local context models to obtain multi-context saliency model using deep convolutional neural networks (CNN). The saliency maps $\text{SM}_i$ obtained from these three methods are then combined to get a single saliency map (see Fig. \ref{Fig-2: Intermideate Results}(c)) as given in equation \ref{eq_2}.
     \begin{equation}
     \label{eq_2}
     \text{SM} = \beta_1 \times \text{SM}_1 + \beta_2 \times \text{SM}_2 + \beta_3 \times \text{SM}_3. 
     \end{equation}
     where, $\beta_1, \beta_2$, and $\beta_3$ are constants. We choose the same value of $\frac{1}{3}$ for $\beta_1, \beta_2$, and $\beta_3$ in this work.
 
\subsection{ Identification of Background and Foreground Superpixels}
\begin{figure}
	\centering
	\stackunder{\epsfig{figure=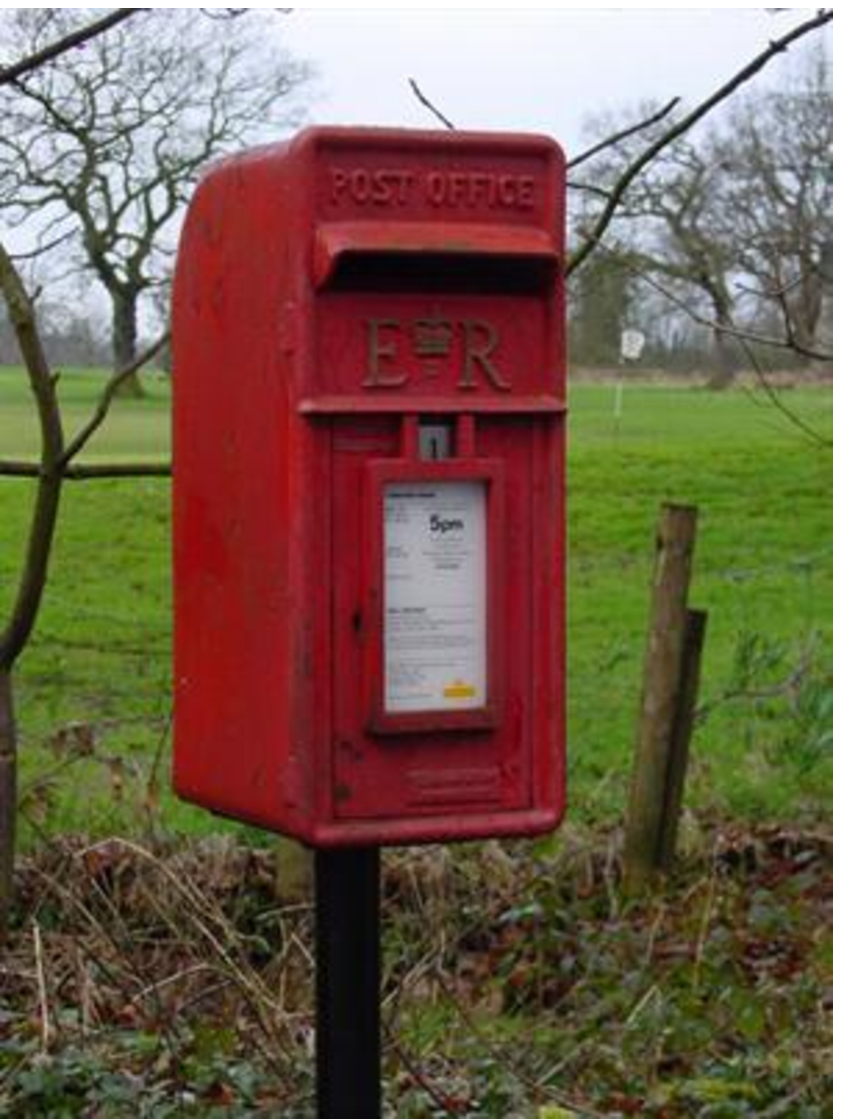,width=0.13\linewidth}}{(a)}
	\stackunder{\epsfig{figure=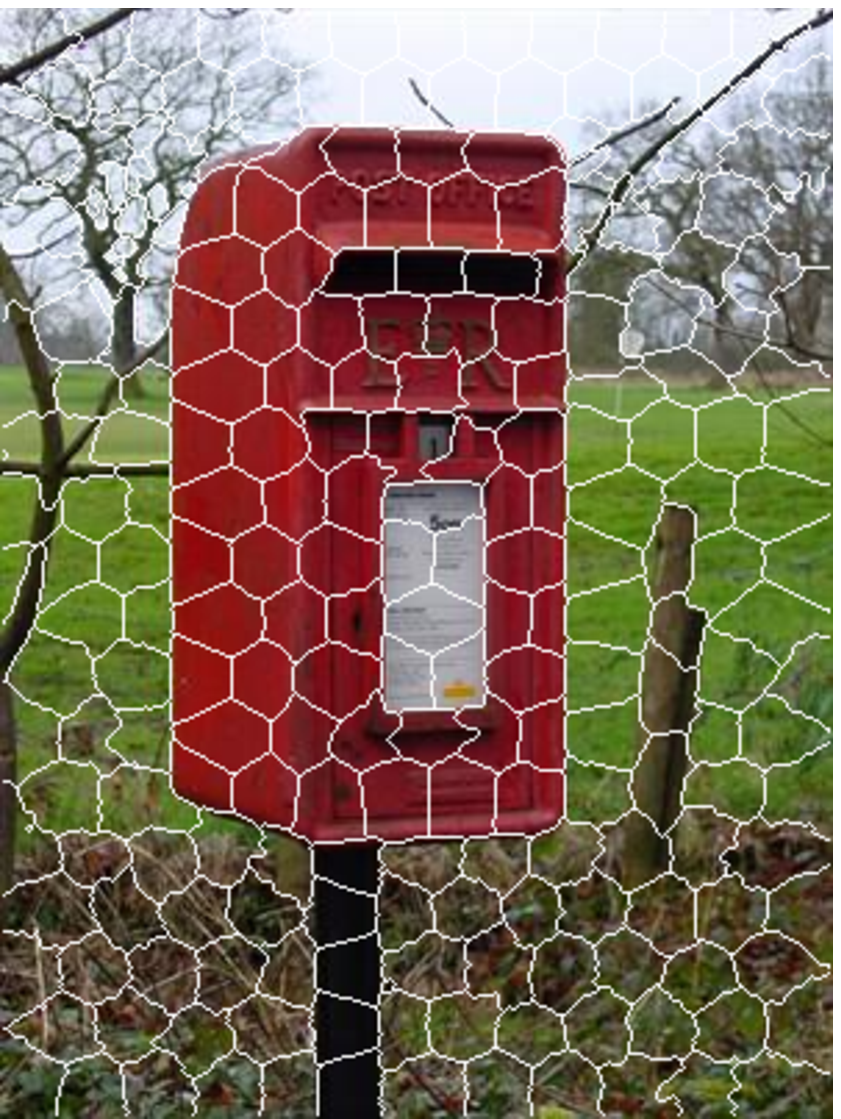,width=0.13\linewidth}}{(b)}
	\stackunder{\epsfig{figure=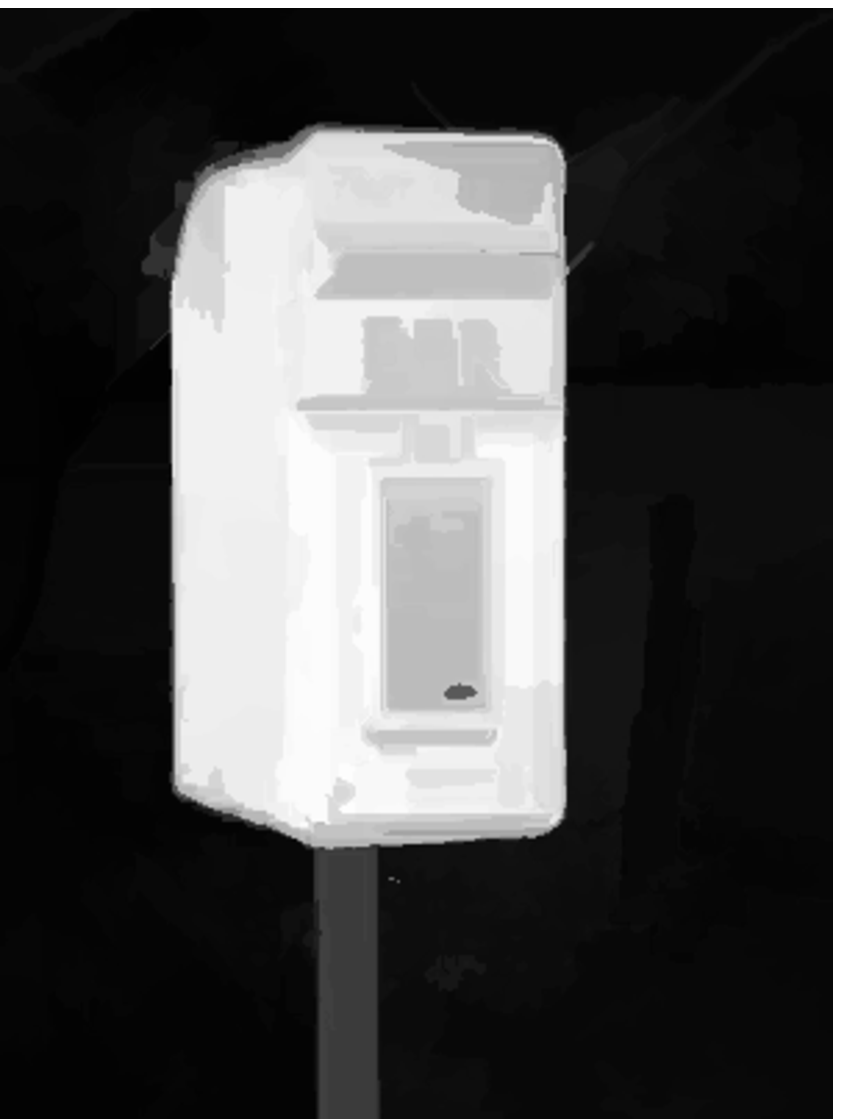,width=0.13\linewidth}}{(c)}
	\stackunder{\epsfig{figure=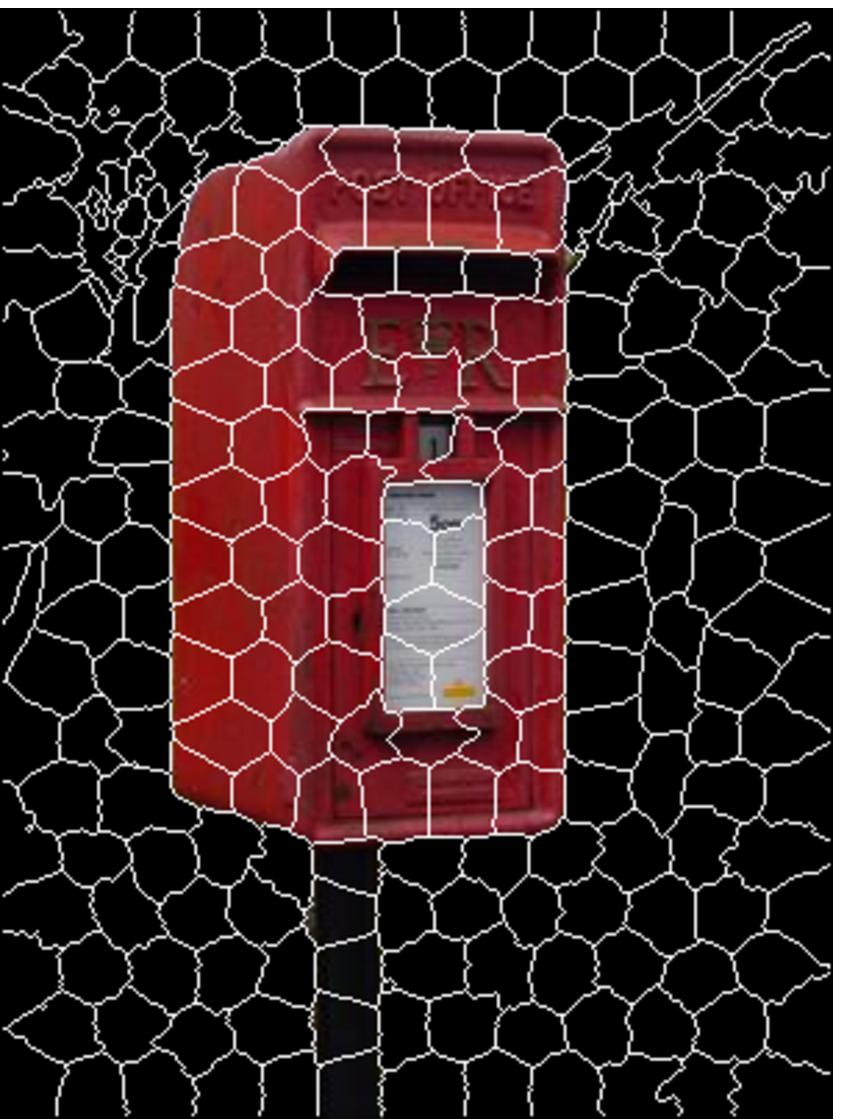,width=0.13\linewidth}}{(d)}
	\stackunder{\epsfig{figure=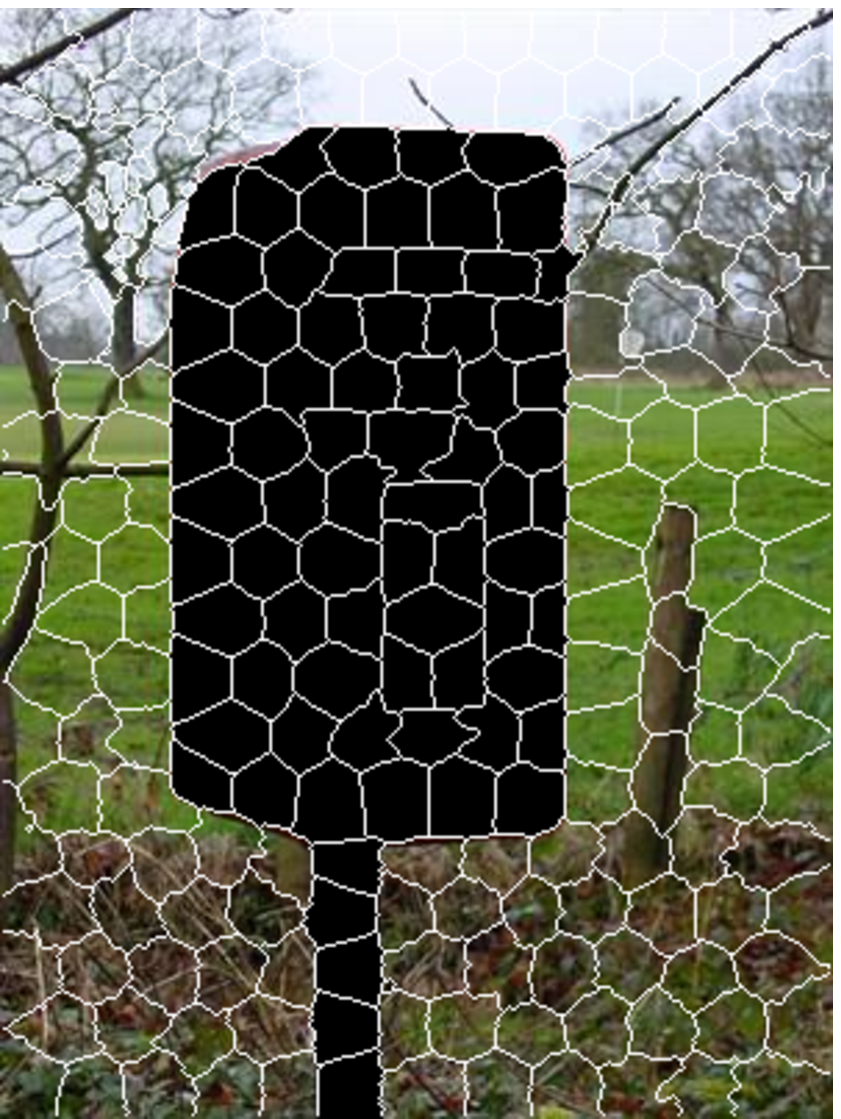,width=0.13\linewidth}}{(e)}
	\stackunder{\epsfig{figure=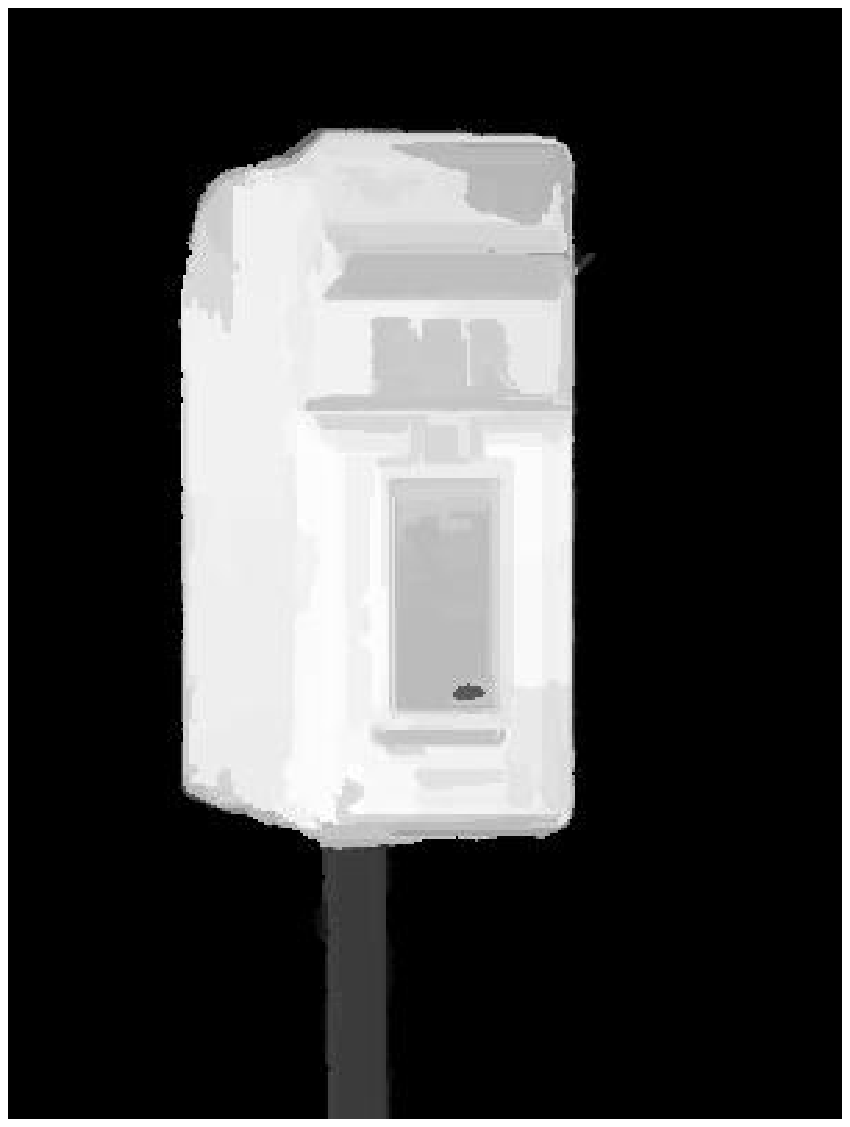,width=0.13\linewidth}}{(f)}
	\stackunder{\epsfig{figure=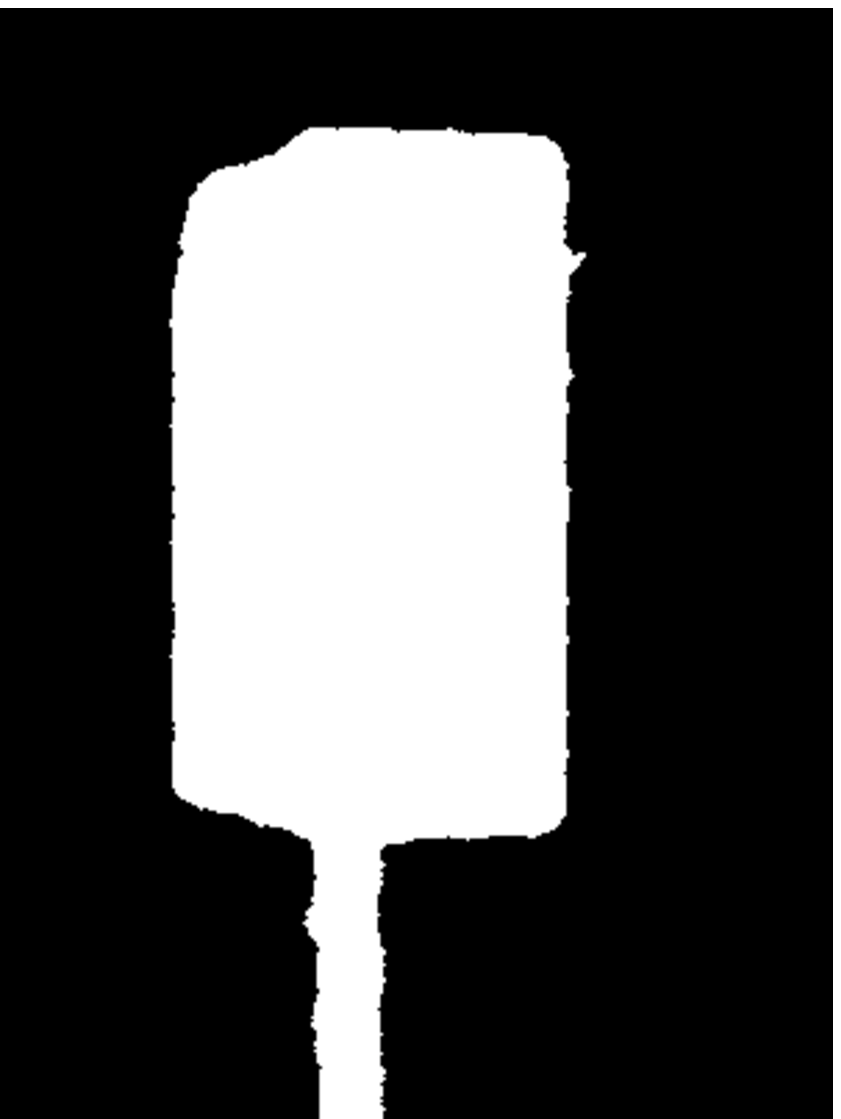,width=0.13\linewidth}}{(g)}
	\stackunder{\epsfig{figure=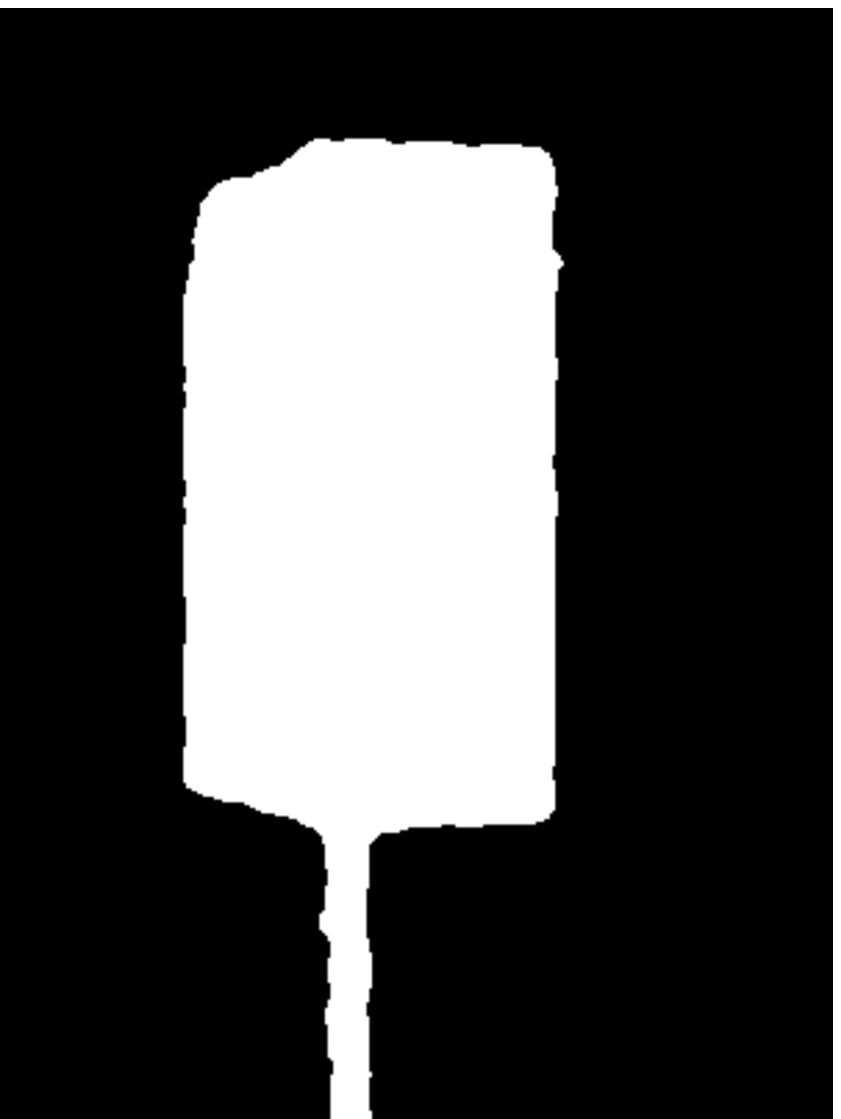,width=0.13\linewidth}}{(h)}
	\stackunder{\epsfig{figure=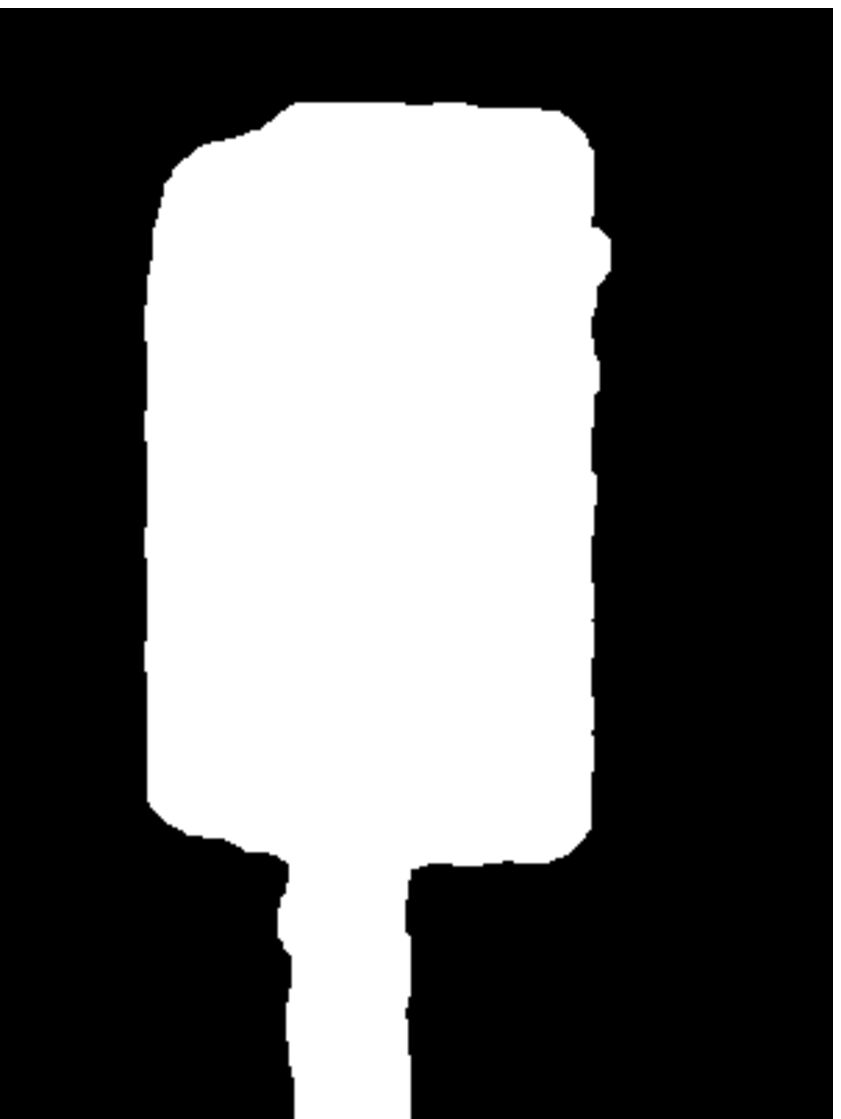,width=0.13\linewidth}}{(i)}
	\stackunder{\epsfig{figure=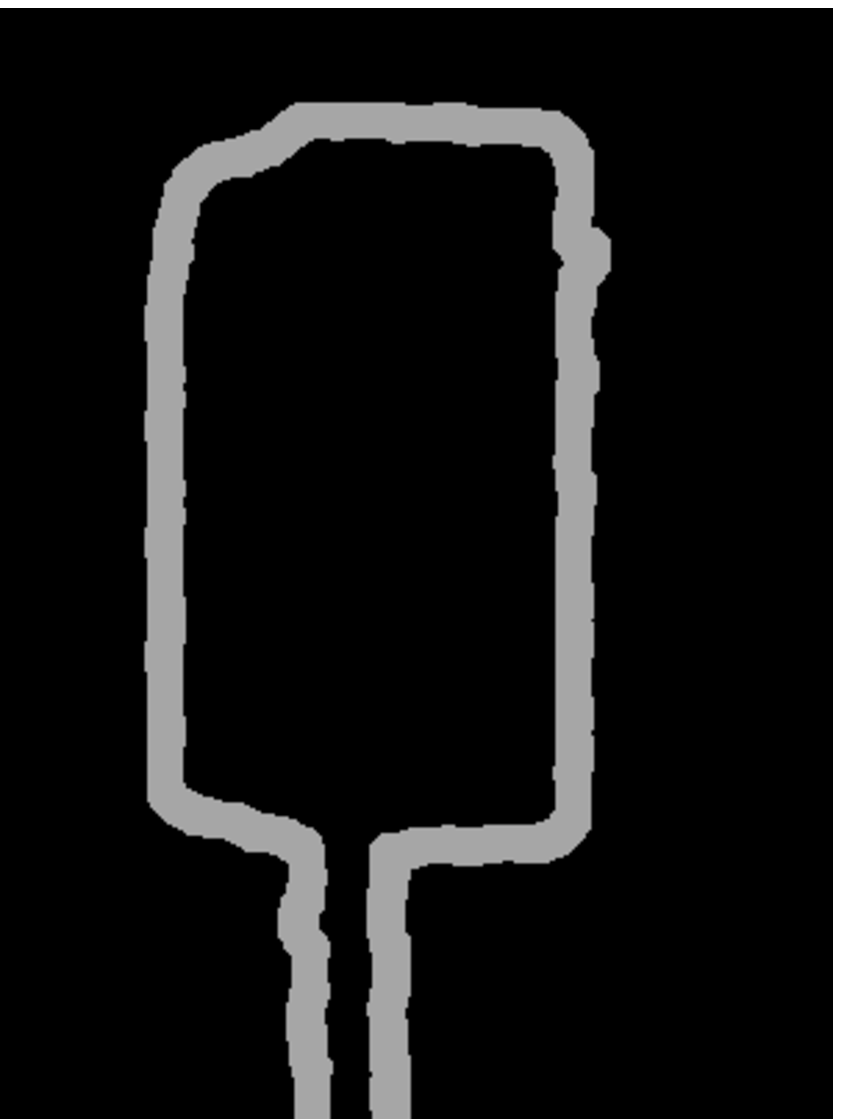,width=0.13\linewidth}}{(j)}
	\stackunder{\epsfig{figure=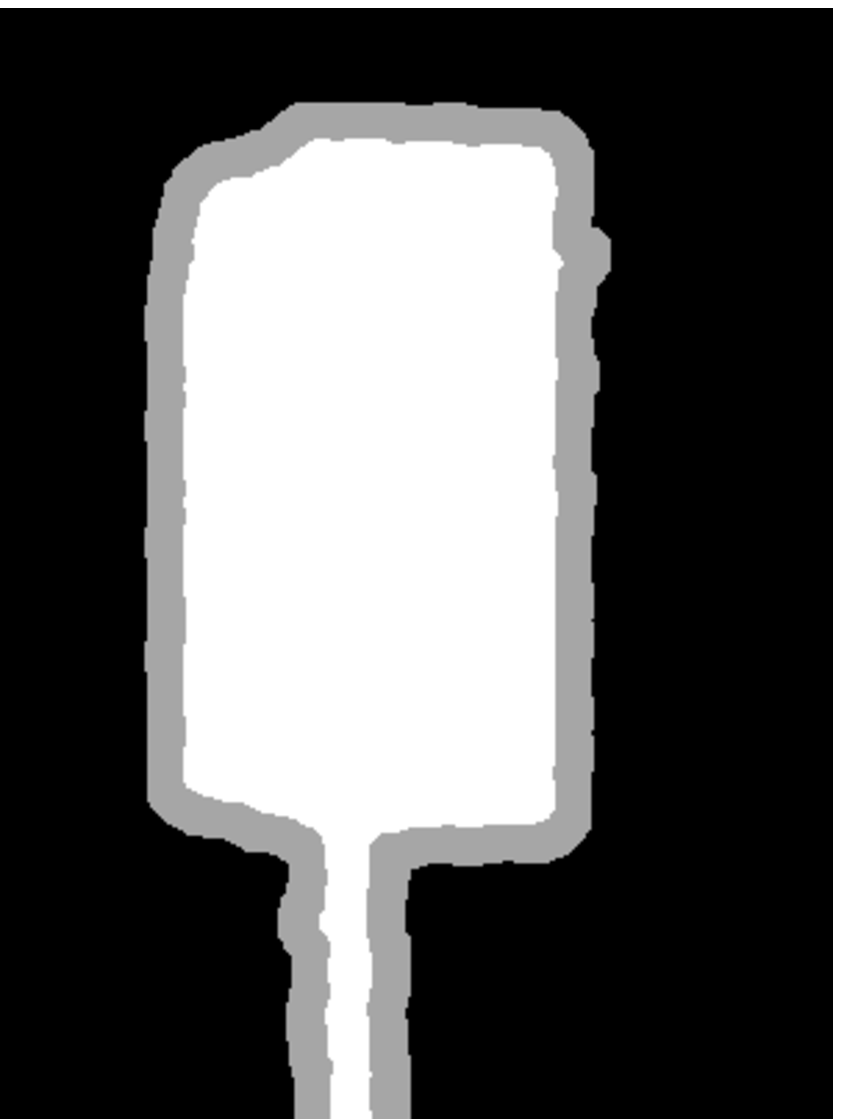,width=0.13\linewidth}}{(k)}
	\stackunder{\epsfig{figure=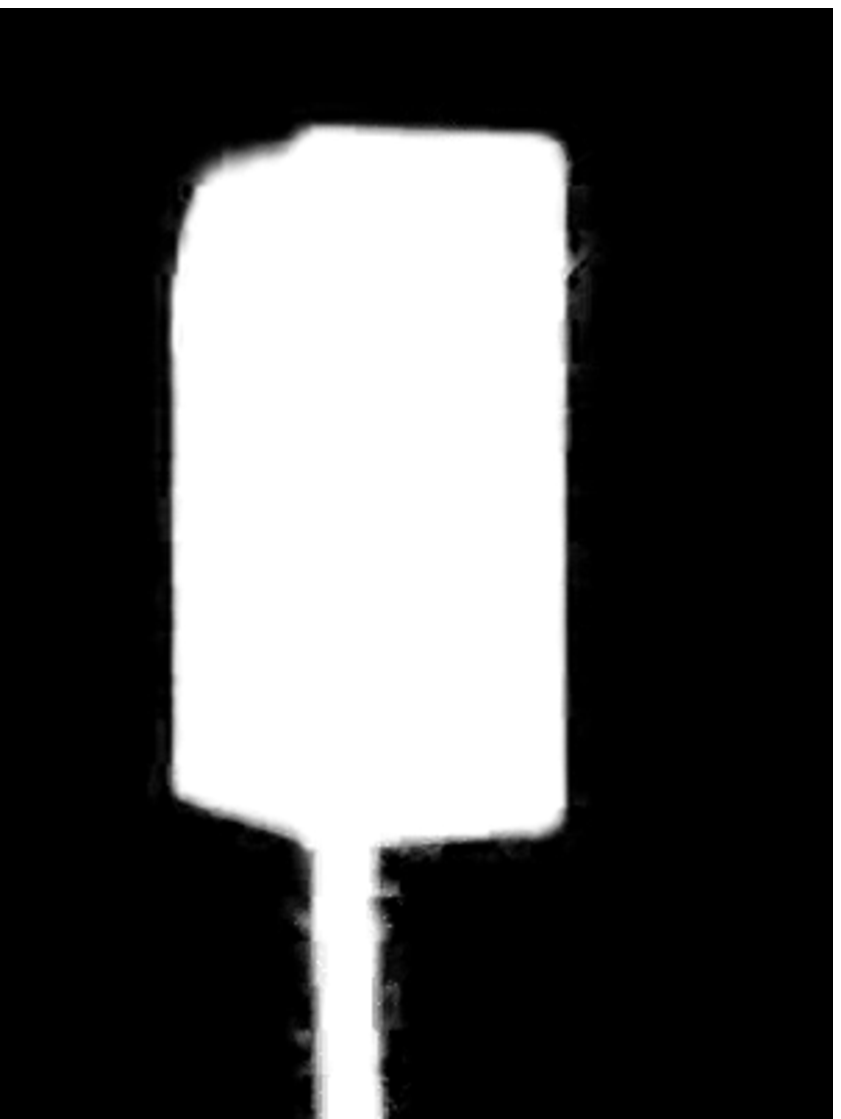,width=0.13\linewidth}}{(l)}
	\caption{Intermediate Results: (a) Input image, (b) Over-segmented image, (c) Saliency map, (d) Foreground superpixel, (e) Background superpixel, (f) Modified saliency map, (g) Binarized saliency map, (h) Eroded saliency map, (i) Dilated saliency map, (j) Differnce, (k) Trimap, (l) Estimated matte using\cite{zheng2009learning}.} 
	\label{Fig-2: Intermideate Results}
\end{figure}

     We use the saliency map $\text{SM}$ to  classify superpixels into salient and non-salient superpixels. For each superpixel, we obtain the median value in the saliency map. If this median value is greater than a threshold $T_1$ then that superpixel is classified as a salient  superpixel.  Otherwise, it is classified as  a non-salient superpixel. Initially, we consider the salient superpixels as foreground superpixels andn non-salient superpixels as background superpixels. It may happen that some salient superpixels belong to background and some non-salient superpixels belong to foreground. To alleviate this problem, we cluster the OTC features of superpixels classified as foreground into five different clusters using $k$-means clustering. Similarly we cluster the OTC features of superpixels classified as background into five different clusters using $k$-means clustering. \\
For each superpixel, which was initially classified as foreground, we compute the euclidean distance $D_{fb}$ between that superpixel and the cluster centers of the background superpixels. If this distance is less than a threshold $T_2$ then that superpixels is identified as a background superpixel. The same process is repeated for the superpixels which were initially classified as background to identify more foreground superpixels.We repeat the same process for all the superpixels identified as background using the cluster center estimated by the foreground superpixels. The separated foreground and background superpixels are shown in Fig. \ref{Fig-2: Intermideate Results}(d, e). Based on this information we modify the saliency map $\text{SM}$ so that only the foreground region will have the salient value. Finally we get the modified saliency map $\text{SM}'$ as shown in Fig. \ref{Fig-2: Intermideate Results}(f).

\subsection{Trimap Generation and Matting}
     To generate the \emph{trimap}, we need a binarized saliency map. The modified saliency map $\text{SM}'$ is binarized using Otsu's thresholding method \cite{otsu1975threshold} as shown in Fig.\ref{Fig-2: Intermideate Results}(g). The binarized saliency map is then eroded and dilated to get the eroded map $\text{SM}_e$  and the dilated map $\text{SM}_d$  as shown in Fig.\ref{Fig-2: Intermideate Results}(h, i). We use a disk structuring element of radii 5 and 10 for the erosion and dilation respectively. The eroded map $\text{SM}_e$ is subtracted from the dilated map $\text{SM}_d$ to get the unknown region of the trimap as given in equation \ref{eq_3}. 
     \begin{equation}
     \label{eq_3}
               \text{SM}_{diff} =  \text{SM}_d - \text{SM}_e
     \end{equation}
     The obtained difference map is multiplied with a constant $C$, where $0<C<1$ (see Fig. \ref{Fig-2: Intermideate Results}(j)). This difference map is then added to the eroded saliency map $\text{SM}_e$, which results into a \emph{trimap} (TM) as shown in Fig. \ref{Fig-2: Intermideate Results}(k). This process is explained in equation \ref{eq_4}. 
\begin{equation}
\label{eq_4}
\text{TM} =  \text{SM}_{diff} + \text{SM}_e
\end{equation}

We use the \emph{Learning based matting}  technique to obtain the alpha matte  for the input image $I$ by using the \emph{trimap} obtained from our proposed framework \cite{zheng2009learning}. The estimated alpha matte is depicted in the Fig. \ref{Fig-2: Intermideate Results}(l).

\section{Results and Discussion}
	\label{Results}
          In this section, we present and discuss the results obtained by our proposed framework. We test our proposed method on a number of images obtained from FT \cite{achanta2009frequency} and PASCAL-S \cite{li2014secrets} datasets. We compare the trimaps generated by the proposed framework with the manually created trimaps. 
 
\begin{figure}
	\centering
	\stackunder{\includegraphics[width=0.19\linewidth]{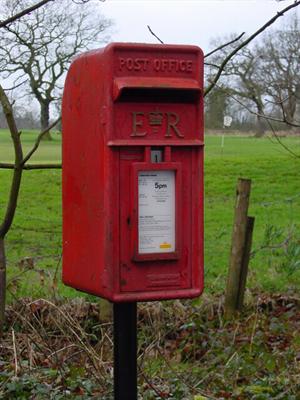}}{}
	\hfill
	\stackunder{\includegraphics[width=0.19\linewidth]{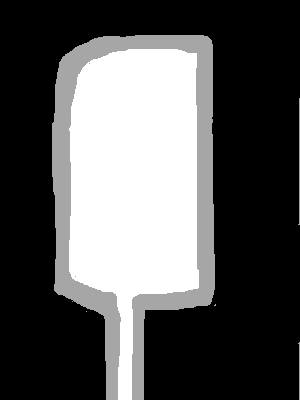}}{}
	\hfill
	\stackunder{\includegraphics[width=0.19\linewidth]{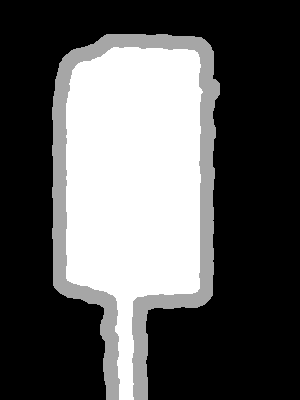}}{}
	\hfill
	\stackunder{\includegraphics[width=0.19\linewidth]{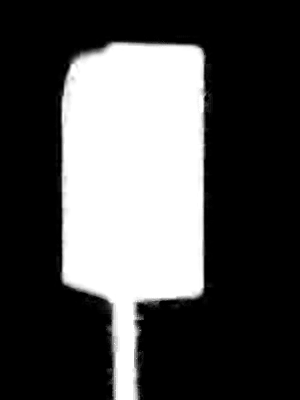}}{}
	\hfill
	\stackunder{\includegraphics[width=0.19\linewidth]{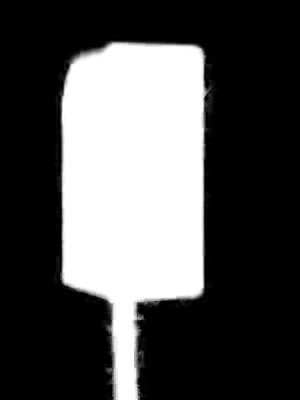}}{}
	\hfill
	\stackunder{\includegraphics[width=0.19\linewidth]{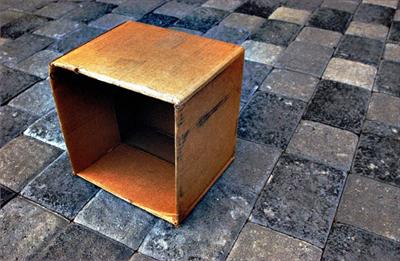}}{}
	\hfill
	\stackunder{\includegraphics[width=0.19\linewidth]{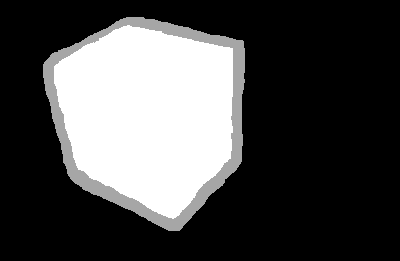}}{}
	\hfill
	\stackunder{\includegraphics[width=0.19\linewidth]{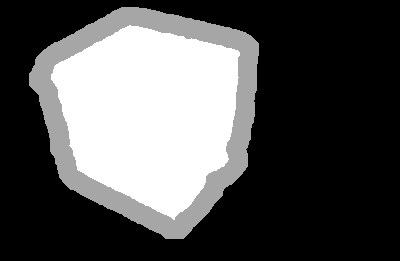}}{}
	\hfill
	\stackunder{\includegraphics[width=0.19\linewidth]{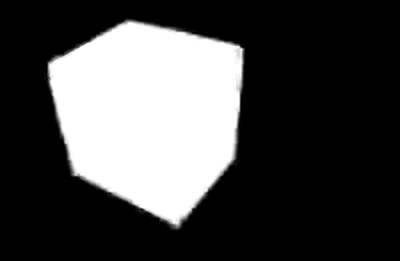}}{}
	\hfill
	\stackunder{\includegraphics[width=0.19\linewidth]{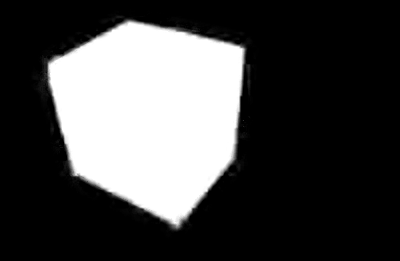}}{}
	\hfill
	\stackunder{\includegraphics[width=0.19\linewidth]{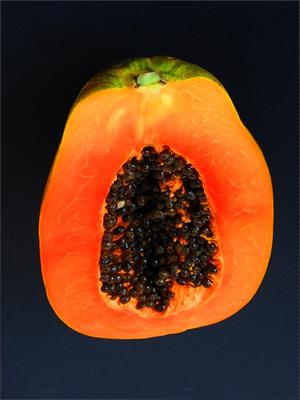}}{}
	\hfill
	\stackunder{\includegraphics[width=0.19\linewidth]{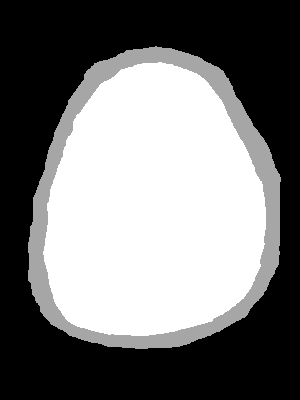}}{}
	\hfill
	\stackunder{\includegraphics[width=0.19\linewidth]{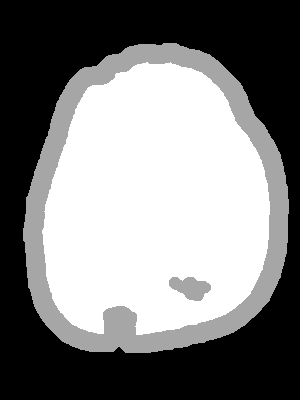}}{}
	\hfill
	\stackunder{\includegraphics[width=0.19\linewidth]{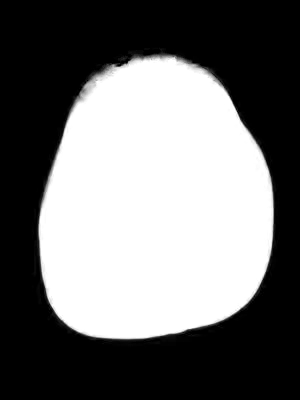}}{}
	\hfill
	\stackunder{\includegraphics[width=0.19\linewidth]{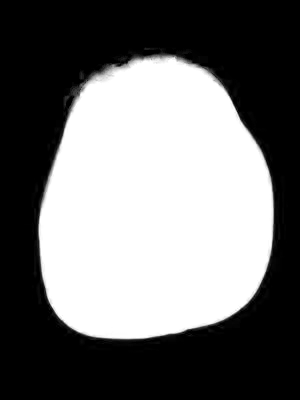}}{}
	\hfill
	\stackunder{\includegraphics[width=0.19\linewidth]{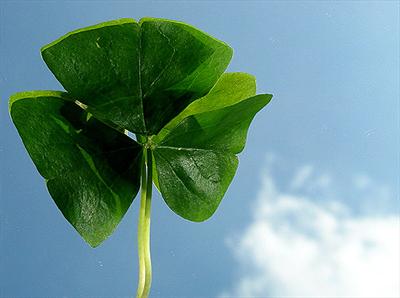}}{}
	\hfill
	\stackunder{\includegraphics[width=0.19\linewidth]{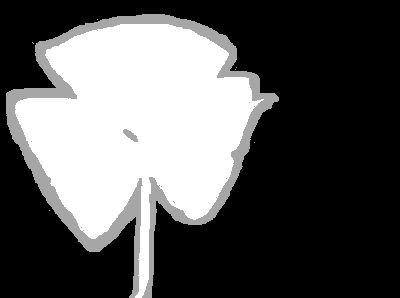}}{}
	\hfill
	\stackunder{\includegraphics[width=0.19\linewidth]{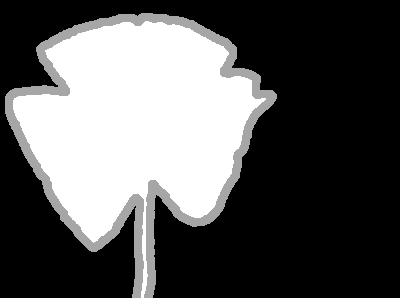}}{}
	\hfill
	\stackunder{\includegraphics[width=0.19\linewidth]{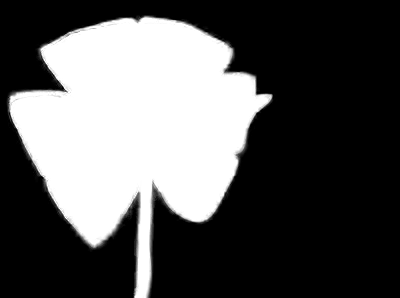}}{}
	\hfill
	\stackunder{\includegraphics[width=0.19\linewidth]{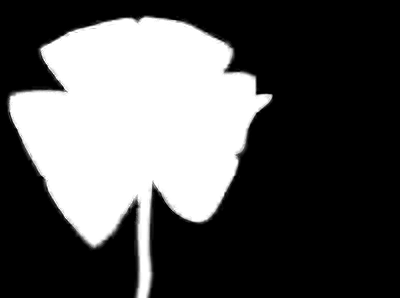}}{}
	\hfill
	\stackunder{\includegraphics[width=0.19\linewidth]{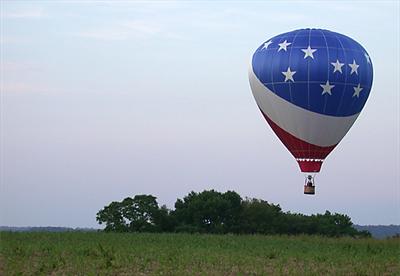}}{}
	\hfill
	\stackunder{\includegraphics[width=0.19\linewidth]{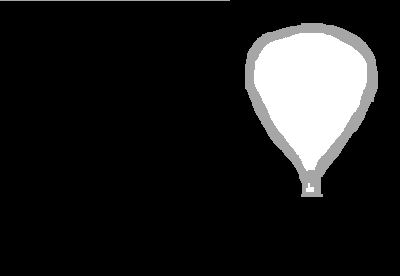}}{}
	\hfill
	\stackunder{\includegraphics[width=0.19\linewidth]{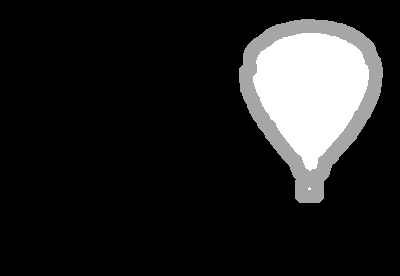}}{}
	\hfill
	\stackunder{\includegraphics[width=0.19\linewidth]{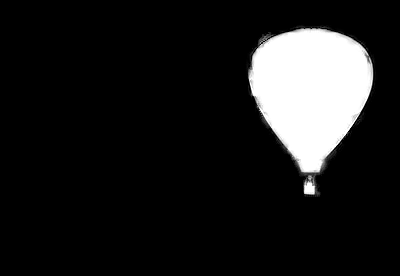}}{}
	\hfill
	\stackunder{\includegraphics[width=0.19\linewidth]{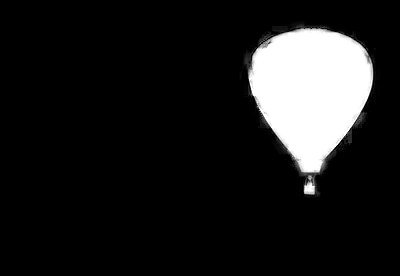}}{}
	\hfill
	\stackunder{\includegraphics[width=0.19\linewidth]{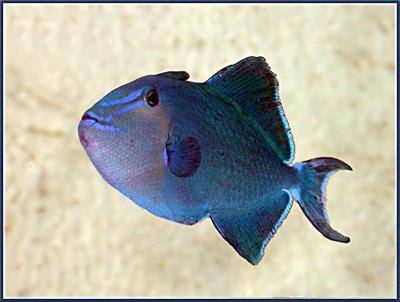}}{}
	\hfill
	\stackunder{\includegraphics[width=0.19\linewidth]{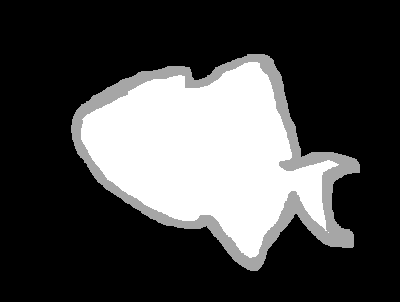}}{}
	\hfill
	\stackunder{\includegraphics[width=0.19\linewidth]{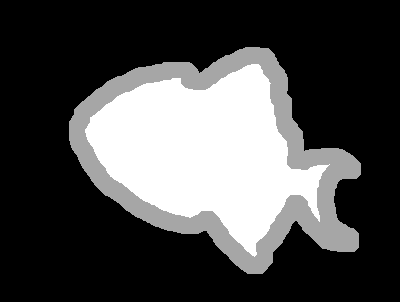}}{}
	\hfill
	\stackunder{\includegraphics[width=0.19\linewidth]{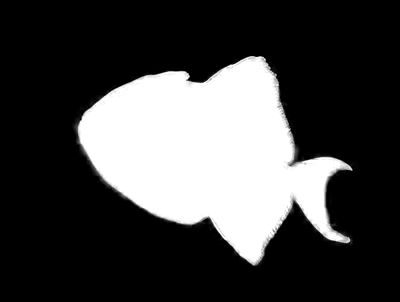}}{}
	\hfill
	\stackunder{\includegraphics[width=0.19\linewidth]{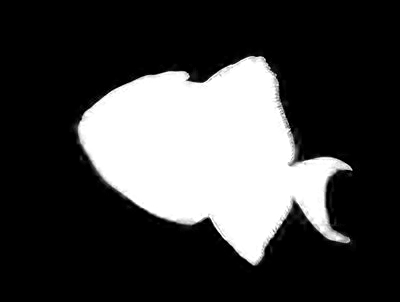}}{}
	\hfill
    \stackunder{\includegraphics[width=0.19\linewidth]{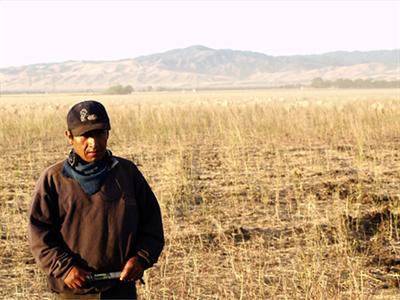}}{(a)}
    \hfill
    \stackunder{\includegraphics[width=0.19\linewidth]{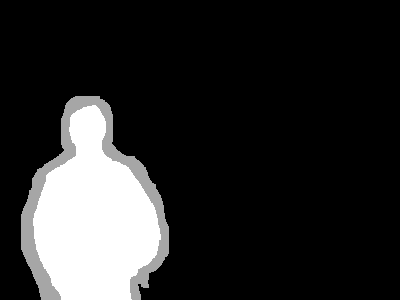}}{(b)}
    \hfill
    \stackunder{\includegraphics[width=0.19\linewidth]{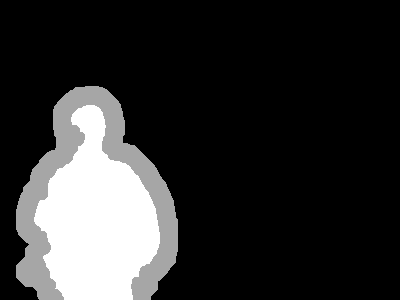}}{(c)}
    \hfill
    \stackunder{\includegraphics[width=0.19\linewidth]{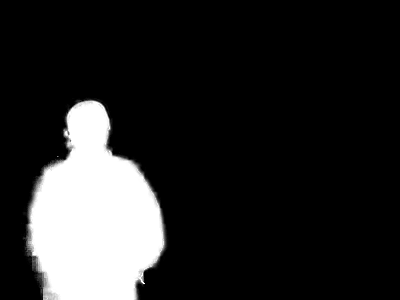}}{(d)}
    \hfill
    \stackunder{\includegraphics[width=0.19\linewidth]{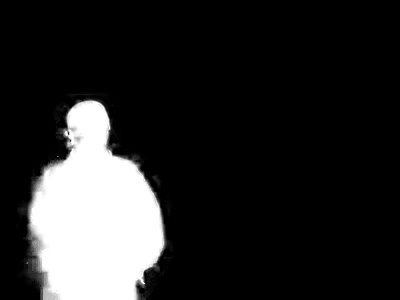}}{(e)}

	\caption{(a) Input image, (b) Trimap (manually generated ), (c) Trimap (Using proposed approach), (d) Matte by using (b) (using \cite{zheng2009learning}), (e) Matte by using (c) (using \cite{zheng2009learning}).} 
	\label{Fig-4:  Results}
\end{figure}

          Our method works well in the case of  images where the background part is natural, which can be noticed in Fig. \ref{Fig-4:  Results}. The first column shows the input images, the second column depicts the manually created trimaps, in the third column the trimaps generated  by the our proposed approach are shown. The mattes corresponding to both these trimaps are shown in the fourth and the fifth column respectively. We employ the matting algorithm proposed in \cite{zheng2009learning}.
          
          The first row of Fig. \ref{Fig-4:  Results} shows the results for an image  which consists of a foreground object (post office box) and a natural background. Here, we can notice that the automatically generated trimap is quite similar to that of manually created trimap thereby leading to accurate matte estimation, similar observation can be made for the images shown in the second, fourth, fifth, and seventh rows. For the image used shown in third row, there is little difference in the automatically  generate trimap and the manually created trimap. Some part of the foreground is marked as unknown in the automatically generated trimap, which is marked as definite foreground in the manually created trimap. However, the matting algorithms takes care of it and we get approximately similar  mattes from both these trimaps. In the sixth row,  we can notice that the trimap obtained using the proposed approach marks the unknown region (foreground boundary) very accurately compared to that of the manually generated trimap.
          
           The results illustrated in Fig. \ref{Fig-4:  Results} demonstrate that the automatically generated trimap is as accurate as the  manually created trimap for generating the mattes. To validate our claim, we  compute   sum of square differences (SSD) for the matte generated using two different trimaps \ie  trimap using the proposed approach  and the manually created trimap. The SSD for the images in the first to the seventh row are  $150, 106, 101, 92, 23, 38, \text{and} \: 58$, respectively. We observed that the SSD values are very small. The proposed method has some limitations which can be observed in the case of  images in which background is synthetically generated. If there is an ambiguity between foreground  and  background color, then the proposed method  might lead to some errors in the trimap. 
                   
            We implemented this framework in MATLAB on a PC with Intel i5-4460s 2.9 GHz processor and 12 GB RAM. For segmenting the image into superpixels we set the value of $N$ in the range of 250 to 400. The threshold $T_1 $ is set to equal to $30 \%$ of the highest salient value in the saliency map. The threshold value $T_2$ is set to equal the mean of  distances between the OTC feature vectors of  superpixels belonging to foreground (or background) and the cluster centers of the background (or foreground) superpixels. The constant C is chosen to be equal to 0.65.  Our proposed method takes only a few seconds to generate the trimap for any given image thereby  automating the entire image matting process.  

%

\section{Conclusion}
\label{conclusion}
Image matting  is an important process for accurate estimation of foreground object from the background in image and video editing applications. This task is ill-posed thereby poses a significant challenge for computational photography. In literature review, we note that almost all the matting algorithms require user intervention in the form of trimap or scribbles as input to these algorithms. The performance of these algorithms depends on these user inputs. Also manually generating a trimap consumes a lot of time. To alleviate this problem and make the whole matting process automatic, we have proposed a simple and efficient framework for automatically generating the trimap for a given input image. The experimental results demonstrate that the automatically generated trimaps are very close to that of manually created trimaps which results in accurate matte estimation. 

We believe  that the automation of the entire matting process will be adapted by researchers and practitioners soon. However, there could be images where there is no distinct salient object presents.  In such a scenario, generating the matte automatically is a challenge to be addressed in future. We would like to extend the proposed approach for processing videos. An automatic  matte generation module is highly desirable for a variety of computational photography task that found by students, researchers, artist, and compositors. We would like to make the approach robust enough so that it would serve as a vital tool for studios in order to generate augmented reality effects in the movies. Another future challenge is to extract mattes corresponding to multiple foreground objects from a background automatically.

	\bibliographystyle{splncs03}

\end{document}